\documentclass{article}

\usepackage{microtype}
\usepackage{graphicx}
\usepackage{subcaption}
\graphicspath{{media/}{src/media/}}
\usepackage{booktabs} 

\usepackage{hyperref}

\usepackage[accepted]{MOSS_camera_ready}

\usepackage{amsmath}
\usepackage{amssymb}
\usepackage{mathtools}
\usepackage{amsthm}

\usepackage[capitalize,noabbrev]{cleveref}

\usepackage{cmd}

\theoremstyle{plain}

\theoremstyle{definition}

\theoremstyle{remark}

\usepackage[textsize=tiny]{todonotes}

\icmltitlerunning{Pruning Increases Orderedness in Weight-Tied Recurrent Computation}

\begin{document}

\onecolumn
\icmltitle{Pruning Increases Orderedness in Weight-Tied Recurrent Computation}



\icmlsetsymbol{equal}{*}

\begin{icmlauthorlist}
\icmlauthor{Yiding Song}{equal,harvard}
\end{icmlauthorlist}

\icmlaffiliation{harvard}{Harvard College, Cambridge, MA, USA}

\icmlcorrespondingauthor{Yiding Song}{yidingsong@college.harvard.edu}

\icmlkeywords{Machine Learning, ICML}

\vskip 0.3in


\printAffiliationsAndNotice{\icmlEqualContribution} 

\begin{abstract}
Inspired by the prevalence of recurrent circuits in biological brains, we investigate the degree to which directionality is a helpful inductive bias for artificial neural networks. Taking directionality as topologically-ordered information flow between neurons, we formalise a perceptron layer with all-to-all connections (mathematically equivalent to a weight-tied recurrent neural network) and demonstrate that directionality, a hallmark of modern feed-forward networks, can be \emph{induced} rather than hard-wired by applying appropriate pruning techniques. Across different random seeds our pruning schemes successfully induce greater topological ordering in information flow between neurons without compromising performance, suggesting that directionality is \emph{not} a prerequisite for learning, but may be an advantageous inductive bias discoverable by gradient descent and sparsification.
\end{abstract}


\section{Introduction}

Neuronal organisation in the biological brain is far less rigid than modern feedforward neural networks. Dynamic synaptic growth and pruning occur early in human Prefrontal Cortex development \citep{kolk2022development}; recurrent computation has been shown to be important in macaque decision-making \citep{mante2013context}; and in contrast to the feedforward architectures of standard vision models, biological vision utilises recurrence extensively \citep{van2020going}.

In light of the lack of inherent directionality in many biological computations, this paper seeks to investigate the necessity of directionality (sometimes referred to in biology as {\it hierarchical organisation}), i.e. that information flow between neurons follows an inherent topological ordering, as a helpful inductive bias within artificial perceptron layers. Starting from an {\it all-to-all} connected architecture where every neuron is connected to every other, we examine whether with the right initialisation and pruning techniques, the resulting network can be finetuned via gradient descent to regain a topological ordering in information flow between neurons.

We make three main contributions within this paper: (i) Formalise the idea of a directionless `complete perceptron' layer similar in form to weight-tied recurrent neural networks; (ii) Introduce a metric to measure topological ordering of information flow in terms of the weight matrix of the complete perceptron layer; and (iii) Investigate various initialisation and pruning methods for inducing such a topological ordering.

\section{Related Work}

Past works have explored how modularity and hierarchical organisation arise within information processing networks. \citet{mengistu2016evolutionary} demonstrated that networks with a connection cost tend to evolve to process information in a hierarchical way. \citet{liu2023seeing} used a similar idea to develop Brain-Inspired Modular Training, showing that regularising by connection cost increases modularity and interpretability in neural networks. However, these works mostly focused on feedforward architectures and did not investigate the impact of connection cost or sparsity on directionality.

Another related body of literature is architecture search via pruning \citep{stothers2019turing,li2022pruning}; to the best of our knowledge, these works focus on feedforward architectures only. The Lottery Ticket Hypothesis \citep{frankle2018lottery,malach2020proving} demonstrated that well-designed pruning can uncover high-performing subnetworks prior to training, motivating our exploration of pruning as an important factor that influences network behaviour in learning.

Architecture-wise, the idea of building models with all-to-all neuronal connections has been found in earlier Boltzmann Machines \citep{ackley1985learning} and Hopfield Networks \citep{hopfield1982neural}; the difference is that our analysis focuses on optimisation via gradient descent and does not minimise an explicit energy function, aligning more closely with traditional Multi-Layer Perceptrons (MLPs) than energy-based networks.

\section{Methods}

\subsection{The Complete Perceptron Layer}
\label{subsec:complete-network}

To empirically verify the benefit of information directionality as an inductive bias, we adapt the standard perceptron layer into a directionless {\it complete perceptron layer} (analogous to the idea of a complete graph), where each neuron is connected to every other neuron within the layer. Formally, a complete perceptron layer consists of $o$ output units, $h$ hidden units, and $i$ input units; we use the convention that the first $o$ neurons are designated as output neurons, the next $h$ hidden neurons, and the final $i$ input neurons. We initialise a weights matrix $\mat{W} \in \R^{(o+h) \times (o+h+i)}$ (the value $\mat{W}_{ij}$ at row $i$ and column $j$ is the strength of the connection from neuron $j$ to neuron $i$), a values vector $\vec{v} \in \R^{(o+h)}$ (initialises the value of each of the $o+h$ output and hidden neurons), and an optional bias vector $\vec{b} \in \R^{(o+h)}$ (the value $\vec{b}_i$ at index $i$ is the bias of the $i$th neuron).

Because a complete perceptron layer is all-to-all connected, in order for its weights to learn complex, non-linear relationships, we run it for multiple iterations, replacing spatial depth with temporal depth. Note that the input neurons are clamped and their values remain unchanged. The computation for a complete perceptron layer is given in \cref{alg:complete}. At a high level, after initialising $\mat{W},\;\vec{v}$, and $\vec{b}$, we evolve the hidden state $\vec{s}$ for a fixed number of iterations $T$, with:
\begin{equation}
    \vec{s}^{(t+1)} = \sigma\br{ [\vec{s}^{(t)}\,\,\vec{x}]\,\mat{W}^{\top} + \vec{b} }, \qquad t=0,\dots,T-1, \label{eq:update}
\end{equation}
where $\vec{s}^{(t)}$ is the hidden state at timestep $t$, $[\cdot\,\,\cdot]$ denotes concatenation along the feature axis (as opposed to the batch axis). Unless otherwise stated, $\sigma$ is the sigmoid function and the initial state $s^{(0)}$ is sampled from a random normal distribution. After $T$ iterations, the layer outputs $\vec{y}=\vec{s}^{(T)}_{[:,\,1:o]}$, the values of the first $o$ neurons along each feature axis.

\subsection{Connections to Existing Architectures}

\cref{eq:update} is in fact equivalent to a recurrent neural network with constant inputs and weights across layers: 
\begin{equation}
    \vec{s}^{(t+1)} = \sigma\br{ \vec{s^{(t)}}\mat{W_s}^{\top} + \vec{x}\mat{W_x}^{\top} + \vec{b} }
\end{equation}
where $\mat{W_s} = \mat{W}_{[:,1:o+h]}$ are the first $o+h$ columns of $\mat{W}$ and $\mat{W_x} = \mat{W}_{[:,o+h+1:o+h+i]}$ are the last $i$ columns of $\mat{W}$, such that $\mat{W} = [\mat{W_s} \mid \mat{W_x}]$. This lends a nice perspective to interpret complete perceptron layers as weight-tied recurrent networks with a constant input, grounding our results in the context of a more commonly studied architecture.

\begin{figure}[t]
\centering
\hfill
\begin{subfigure}{0.49\textwidth}
    \centering
    \includegraphics[width=\textwidth]{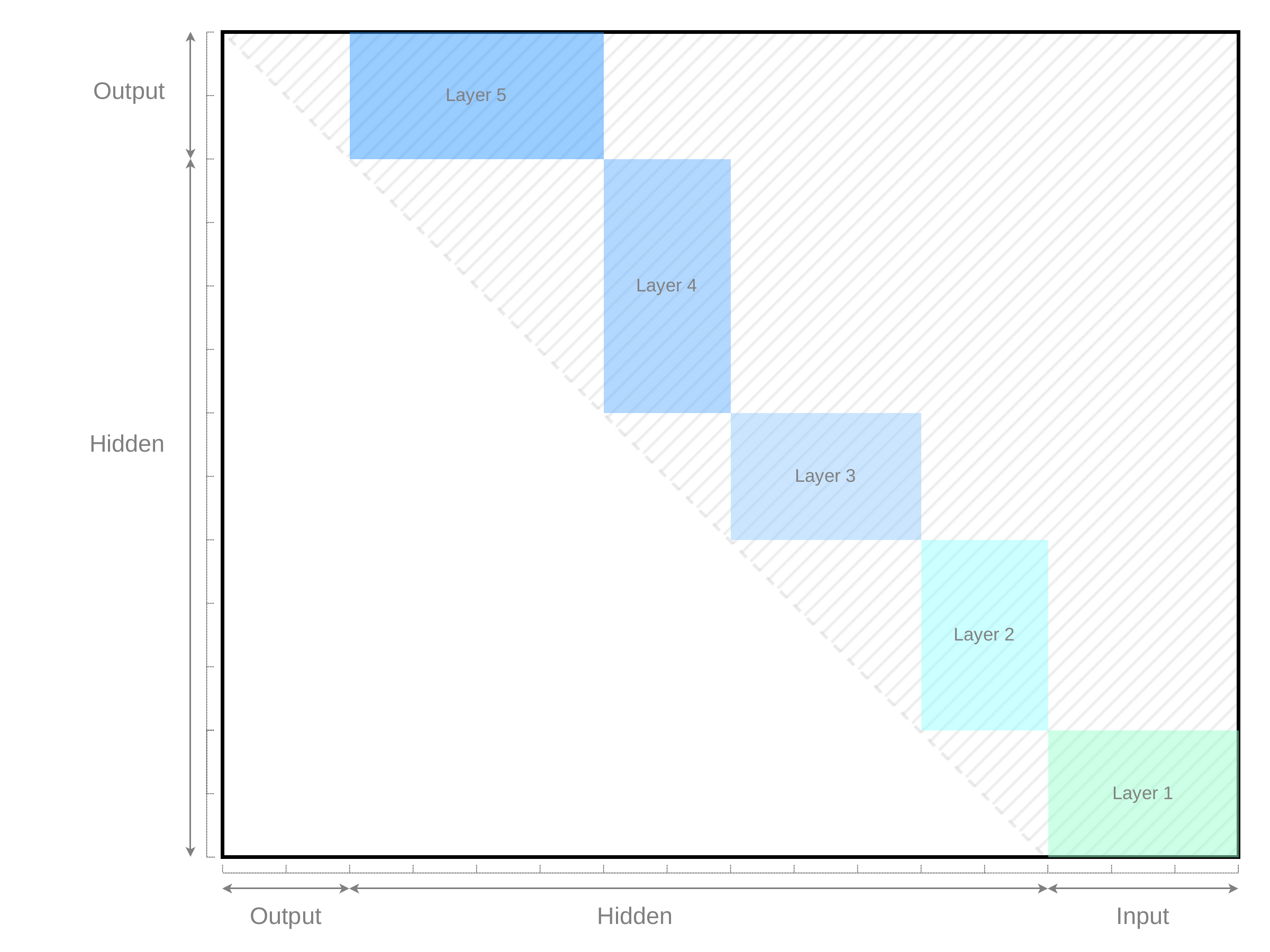}
    \caption{Weights perfectly simulating a normal multilayer perceptron.}
    \label{fig:render-mlp}
\end{subfigure}
\hfill
\begin{subfigure}{0.49\textwidth}
    \centering
    \includegraphics[width=\textwidth]{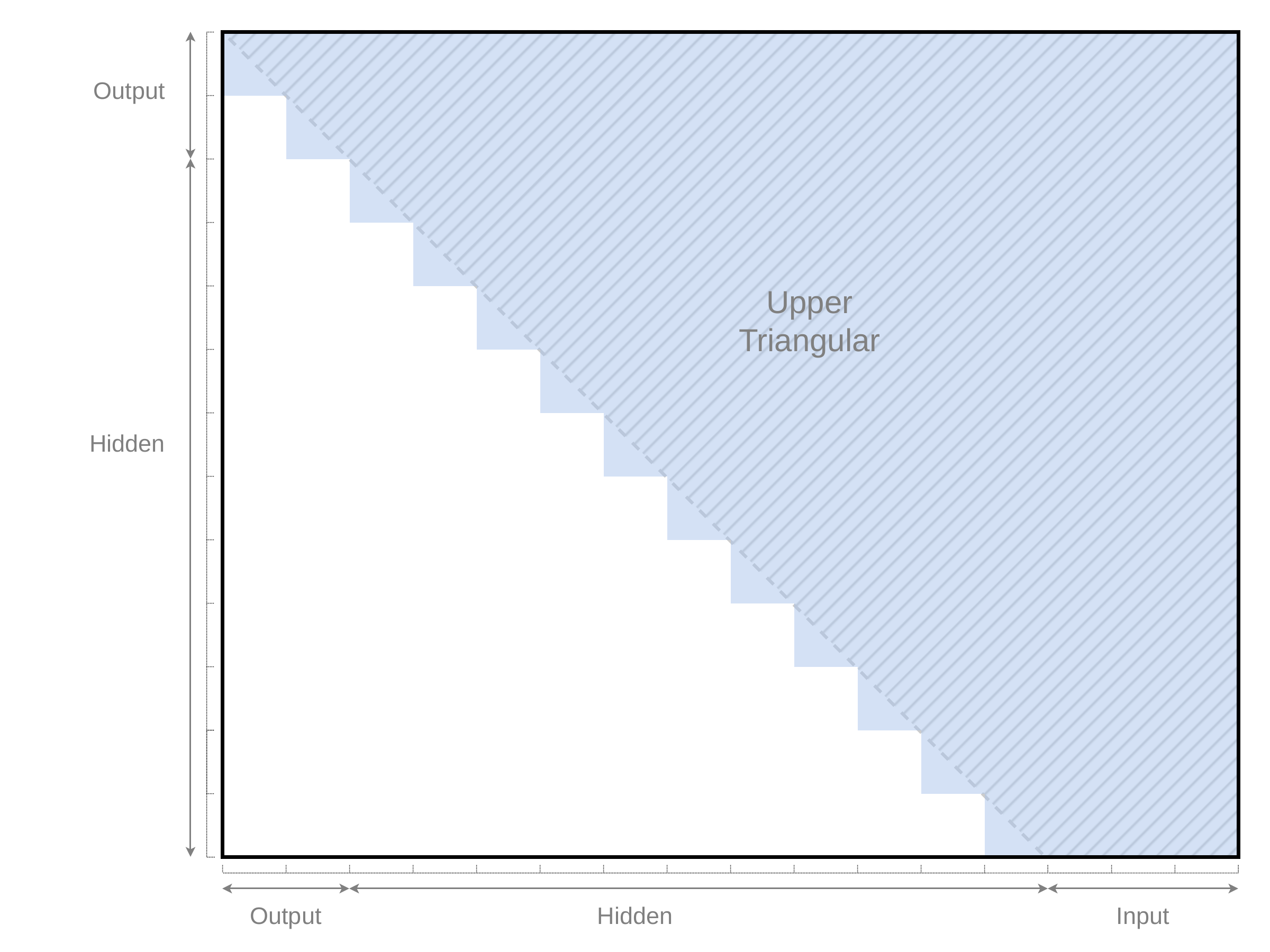}
    \caption{Weights with perfect topological ordering.}
    \label{fig:render-topological}
\end{subfigure}
\hfill
\caption{Visualisation of the weights matrix of a complete perceptron layer for two different cases: perfect simulation of a standard multi-layer perceptron and perfect topological ordering.}
\label{fig:renders}
\end{figure}

Nevertheless, we choose to define the network using \cref{eq:update} because it is easier to interpret for the questions posed in this paper. Note that at each timestep, each hidden unit $i$ performs a weighted sum (followed by some activation function) over its inputs, with the weights defined by row $i$ in the weights matrix $\mat{W}$. Therefore, if the weights of a complete perceptron layer were to simulate a perfectly feedforward MLP when executed, then $\mat{W}$ would consist of rectangular blocks above the main diagonal (see \cref{fig:render-mlp}). Each block corresponds to the weights of one layer in the MLP attending to the outputs of the previous layer; the first layer only attends to the inputs. In the case of \cref{fig:render-mlp}, if the complete perceptron layer were to be computed over 5
iterations, it would simulate the 5-layer MLP whose layer weights are contained in each block.

No such block structure emerged from the gradient descent and pruning methods experimented in this paper. Therefore we generalise the notion of layer-wise feedforward directionality to a {\it topological ordering} of information flow between neurons: that is, earlier neurons only feed information into later neurons, and information does not flow back. On the weights matrix, this corresponds to an upper triangular matrix,\footnote{Here we will allow values on the diagonal, as we choose to permit a neuron to communicate with itself.} because later neurons are allowed to attend to the outputs of earlier ones, but not vice versa. Such a weights matrix shares a similar pattern of information flow to an MLP with only one neuron in each layer and many weighted, feedforward residual connections.

\subsection{Orderedness}

Because the indexing of the hidden units within a weights matrix is arbitrary, we cannot simply measure the `orderedness' of a weights matrix by comparing it with its upper-triangular equivalent; we must take into account all the possible indexing of the hidden units. Therefore, to establish a good metric for the {\it orderedness} of a network, we wish to find the permutation of hidden units such that least of the weights are included in the lower triangular matrix. This reduces `back-flow' of information and minimises the weights that would need to be discarded to achieve perfect topological ordering.

More formally, denote by $\Pi$ the set of all permutations on the order of the hidden neurons on the weight matrix $\mat{W}$. Define by $L(\mat{W})$ the sum of the lower-triangular values in matrix $\mat{W}$ (excluding the diagonal), and by $S(\mat{W})$ the sum of all the values in matrix $\mat{W}$. Additionally, let $\mat{W}^{\mathrm{abs}}$ take the absolute value of all elements in matrix $\mat{W}$. Then, the {\it orderedness} $O$ of the weight matrix $\mat{W}$ is defined as:
\begin{equation}
    O(\mat{W}) = 1 - \br{ \min_{\pi \in \Pi} \frac{L(\pi(\mat{W}^{\mathrm{abs}}))}{S(\pi(\mat{W}^{\mathrm{abs}}))} }
\end{equation}

The metric $O(\mat{W})$ measures the proportion of weights that flow forward in the complete perceptron layer when the hidden units are organised in such as way as to minimise feedback edges. This quantity is 1 in the limit of a perfectly topologically-ordered feedforward network.

\subsection{Initialisation and Pruning}

We evaluate the emergence of orderedness under a set of initialisation schemes for the weights and values, as well as a suite of pruning operations that act \emph{directly} on the weight matrix $W$ during training. The full quantitative effects are reported in Table~\ref{tab:quantitative}; the details of each method are explained in the Appendix (\cref{sec:init-prune}).

Empirically, we found that simply training the complete perceptron layer on synthetic tasks is not sufficient to induce greater orderedness. Inspired by the work of \citet{mengistu2016evolutionary,liu2023seeing} investigating the relationship between connection cost and modularity, we were motivated to see whether enforcing sparsity could encourage the model to be more economical with its high-magnitude weights, and whether that would lead to increased orderedness for efficient computation.

\section{Experiments}

Because of the limitations of compute and the low scalability of the complete perceptron layer, we chose to use very simple tasks for our experiments. Nevertheless, the point of interest here is the learning dynamics of these complete networks. We used two datasets for our experiments: XOR and Sine. For the XOR task, the input is two binary numbers and the output is the binary XOR of the two inputs. For the Sine task, the input is two real numbers $a, b, \in [0, 3]$ and the output is $(\sin(a) + \sin(b)) / 2$ (we chose the range so that all outputs are positive, in order to suit the Sigmoid activation function).

The standard hyperparameters used for the complete perceptron layer in the default control setup are presented in the Appendix (\cref{sec:hyperparam}). To investigate the emergence of orderedness beyond the default control setup, we varied the way we initialised values and weights and experimented with different pruning methods. Additionally, we tried to establish a relationship between the number of hidden units within the complete layer, the number of iterations we evolve it for, and the resulting orderedness of the weights. For pruning, we also investigated how the desired sparsity levels affected orderedness.

\section{Results and Discussion}

A graph of the training losses of the complete perceptron layer on both tasks is shown in Appendix (\cref{sec:training}), presented in comparison to the training losses of traditional MLPs. In both cases, loss descends smoothly despite the absence of pre-imposed directionality, establishing the proof-of-concept that a complete perceptron architecture is trainable. 

\begin{table}[!t]
\centering
\caption{Quantitative effects of various initialisation and pruning methods on orderedness. All data is reported averaged across 10 seeds, with standard deviation after the $\pm$ sign.}
\label{tab:quantitative}
\begin{subtable}[c]{\textwidth}
\centering
\subcaption{Effect of various initialisation methods on the change in orderedness pre- and post-training ($\Delta O$). By default, random normal initialisation is applied on the weights matrix and values vector. Initial orderedness pre-training ($O$) is provided for context.}
\label{tab:initialisation}
\begin{tabular}{|r|r|r|r|r|}
    \hline
    \textbf{Initialisation} & \textbf{$O$ (Untrained)} & \textbf{$\Delta O$ (XOR)} & \textbf{$\Delta O$ (Sine)} \\ \hline
    {\it Default} &
        0.672 $\pm$ 0.050 &
        0.032 $\pm$ 0.047 &
        0.023 $\pm$ 0.033 \\
    Random Uniform ($\mat{W}$) &
        0.647 $\pm$ 0.045 &
        0.075 $\pm$ 0.057 &
        0.076 $\pm$ 0.039 \\
    Zeros ($\vec{v}$) &
        0.680 $\pm$ 0.070 &
        0.049 $\pm$ 0.078 &
        -0.004 $\pm$ 0.030 \\
    \hline
\end{tabular}
\end{subtable}

\vspace{0.3cm}

\begin{subtable}[c]{\textwidth}
\centering
\subcaption{Effect of various pruning methods on change in orderedness pre- and post-training ($\Delta O$). The change in orderedness without training is provided as a control. We prune the weights matrix only.}
\label{tab:pruning}
\begin{tabular}{|r|r|r|r|}
    \hline
    \textbf{Pruning} (on $\mat{W}$) & \textbf{$\Delta O$ (Untrained)} & \textbf{$\Delta O$ (XOR)} & \textbf{$\Delta O$ (Sine)} \\ \hline
    {\it Default}: none &
        0.000 $\pm$ 0.000 &
        0.032 $\pm$ 0.047 &
        0.023 $\pm$ 0.033 \\
    Random ($p$=0.5) &
        0.170 $\pm$ 0.062 &
        0.243 $\pm$ 0.082 &
        0.104 $\pm$ 0.047 \\
    Top-K ($k$=0.5) &
        0.039 $\pm$ 0.032 &
        0.116 $\pm$ 0.060 &
        0.086 $\pm$ 0.037 \\
    Dyn. Top-K ($k$=0.5) &
        0.038 $\pm$ 0.025 &
        0.149 $\pm$ 0.073 &
        0.086 $\pm$ 0.037 \\
    Tril-damp ($f$=0.8) &
        0.328 $\pm$ 0.050 &
        0.280 $\pm$ 0.045 &
        0.343 $\pm$ 0.026 \\
    Dyn. Tril-damp ($f$=0.8) &
        0.315 $\pm$ 0.048 &
        0.280 $\pm$ 0.045 &
        0.343 $\pm$ 0.026 \\
    \hline
\end{tabular}
\end{subtable}
\end{table}

\paragraph{Pruning Correlates with Orderedness.} \cref{tab:quantitative} displays the resulting orderedness of different initialisation and pruning techniques. Initialisation does not appear to play a huge role in increasing orderedness, whilst various pruning methods see more success. It is expected that Tril-Damp methods (which encourage an upper-triangular weights matrix) lead to a large increase in the orderedness of networks with their explicit inductive bias. More surprising is that the two variants of Top-K pruning, which select for weights with the largest absolute value and are agnostic to orderedness, lead to an increase in orderedness without the explicit need for any other inductive bias. The much lower distribution of values of $\Delta O$ from an untrained network indicates that this observation is unlikely to be due to the nature of Top-K sparsification itself, suggesting that under correct pruning techniques, orderedness can be induced naturally. Random pruning works surprisingly well, but this is more so an artefact of the mathematical definition of orderedness, because even on an untrained network, random pruning leads to an increase in orderedness.

\begin{figure}[t]
\centering
\begin{subfigure}{0.49\textwidth}
    \centering
    \includegraphics[width=\textwidth]{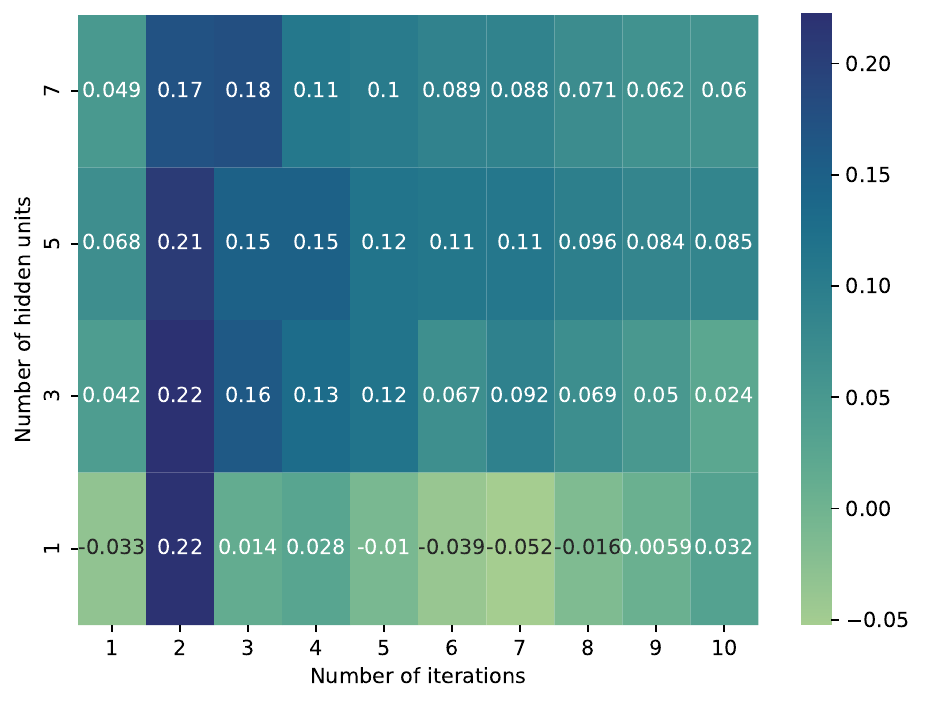}
    \caption{Mean $\Delta O$ for the XOR task.}
\end{subfigure}
\hfill
\begin{subfigure}{0.49\textwidth}
    \centering
    \includegraphics[width=\textwidth]{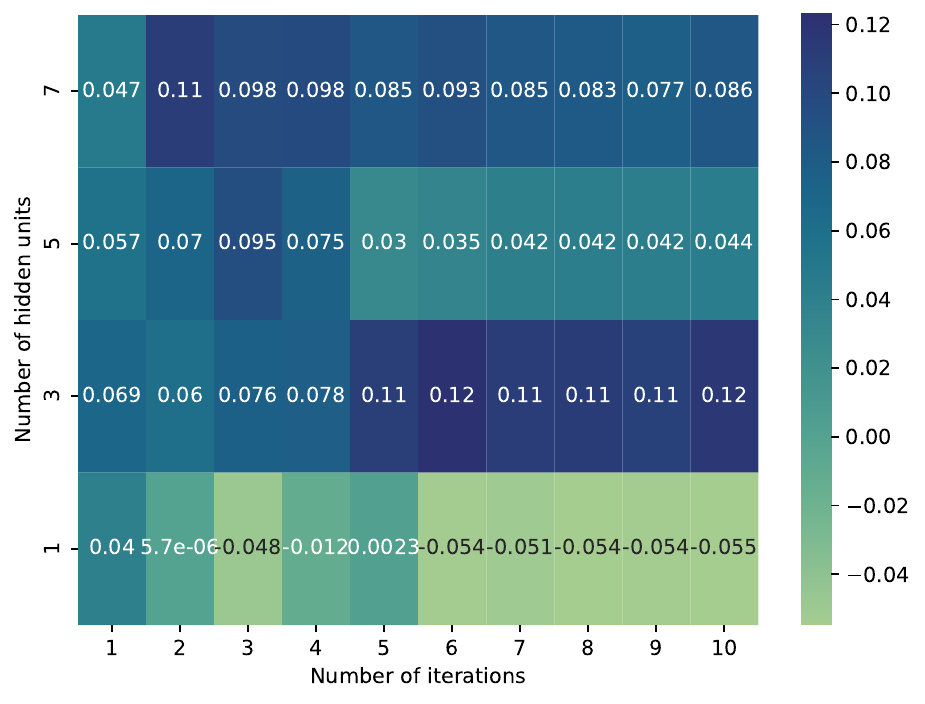}
    \caption{Mean $\Delta O$ for the Sine task.}
\end{subfigure}
\caption{Change in orderedness of the weights matrix after training ($\Delta O$) as a function of both hidden units and number of iterations for the XOR and Sine tasks. Here $\Delta O$ is reported as the mean across 10 seeds. Dynamic Top-K (with $k=0.5$) was used for pruning.}
\label{fig:hi-plot}
\end{figure}

\begin{figure}[!t]
\centering
\hfill
\begin{subfigure}{0.33\textwidth}
    \centering
    \includegraphics[width=\textwidth]{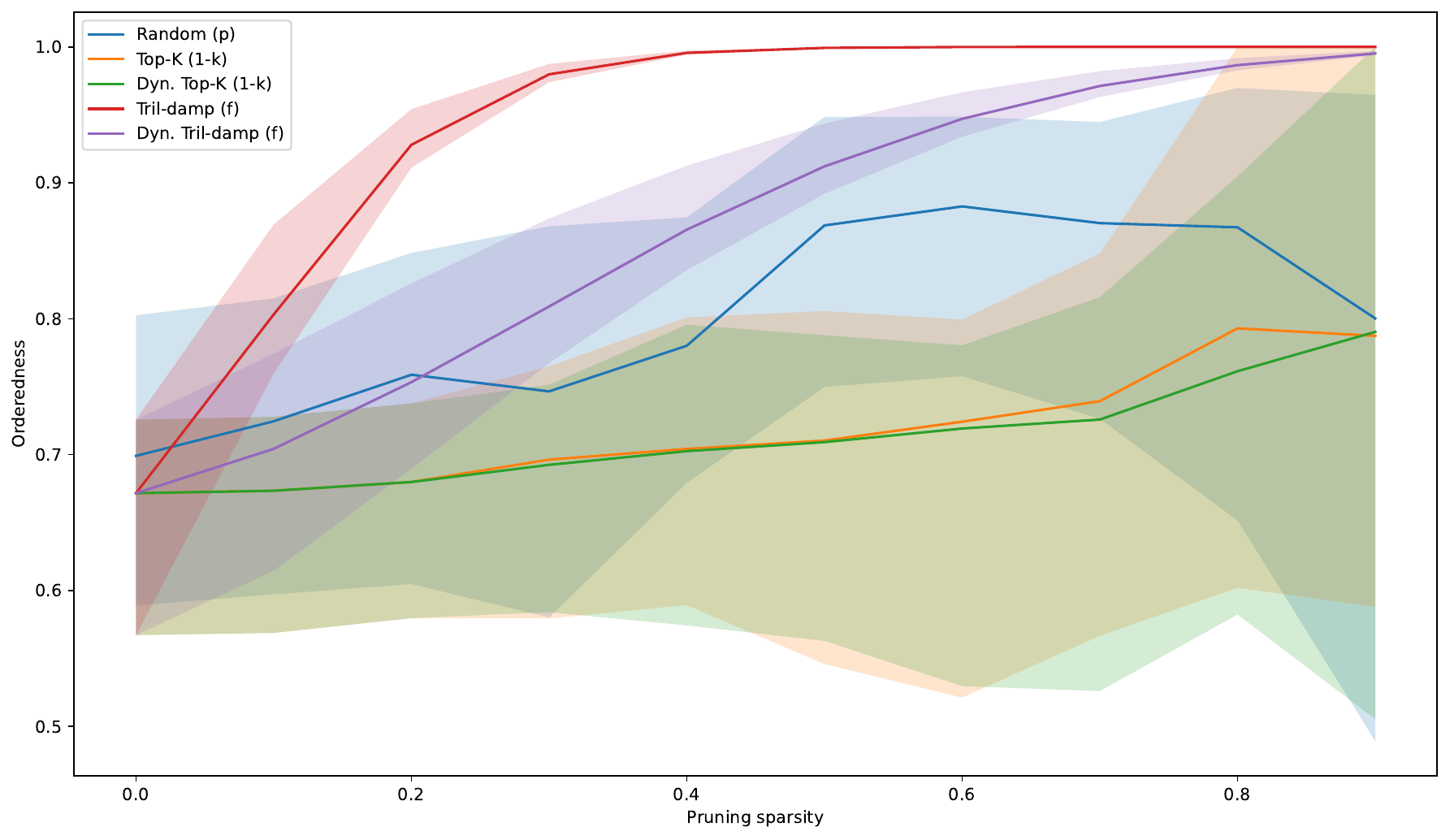}
    \caption{Relationship between pruning sparsity \\ and orderedness (untrained).}
    \label{fig:so-none}
\end{subfigure}
\hfill
\begin{subfigure}{0.33\textwidth}
    \centering
    \includegraphics[width=\textwidth]{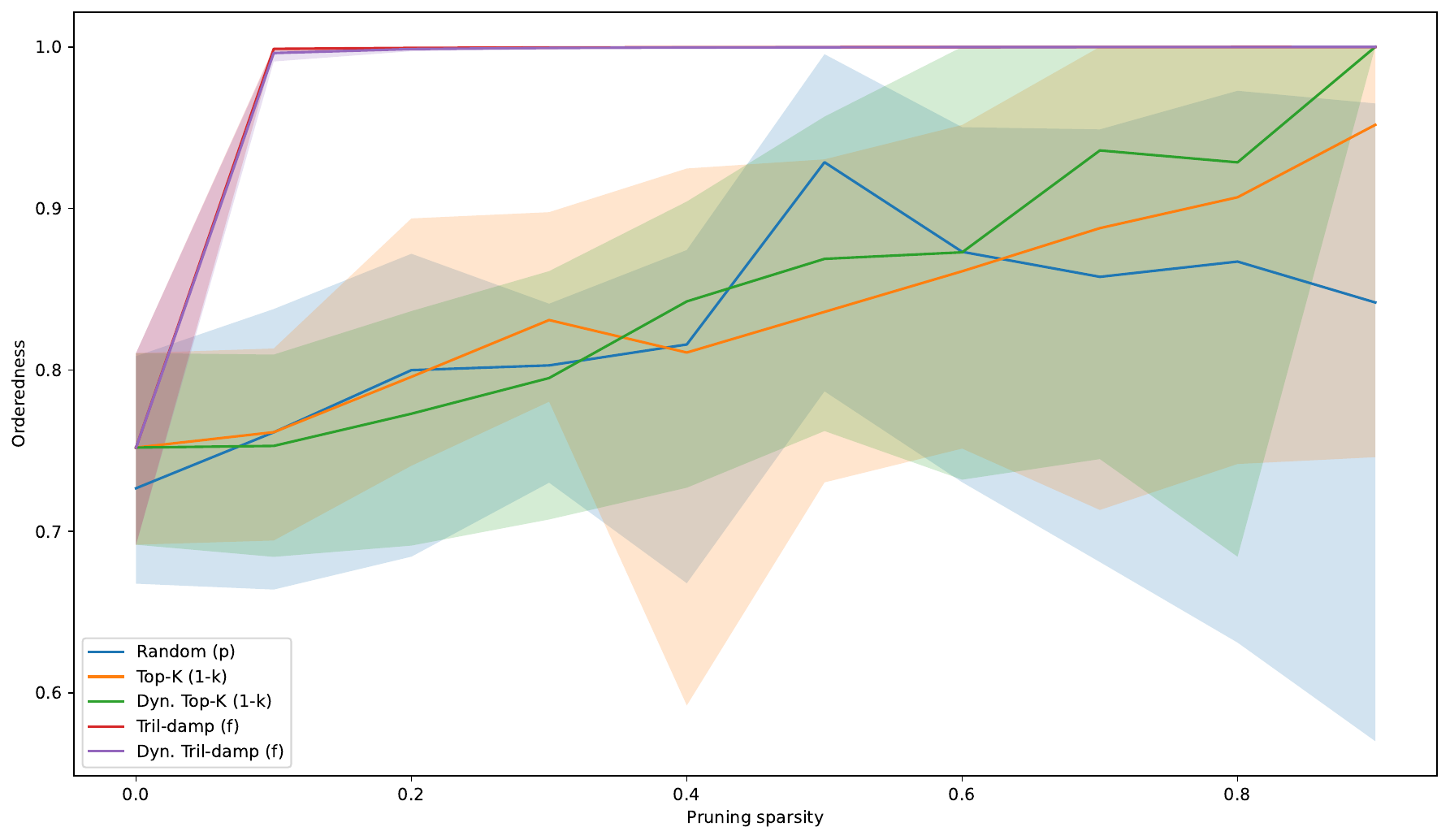}
    \caption{Relationship between pruning sparsity \\ and orderedness (XOR).}
\end{subfigure}
\hfill
\begin{subfigure}{0.33\textwidth}
    \centering
    \includegraphics[width=\textwidth]{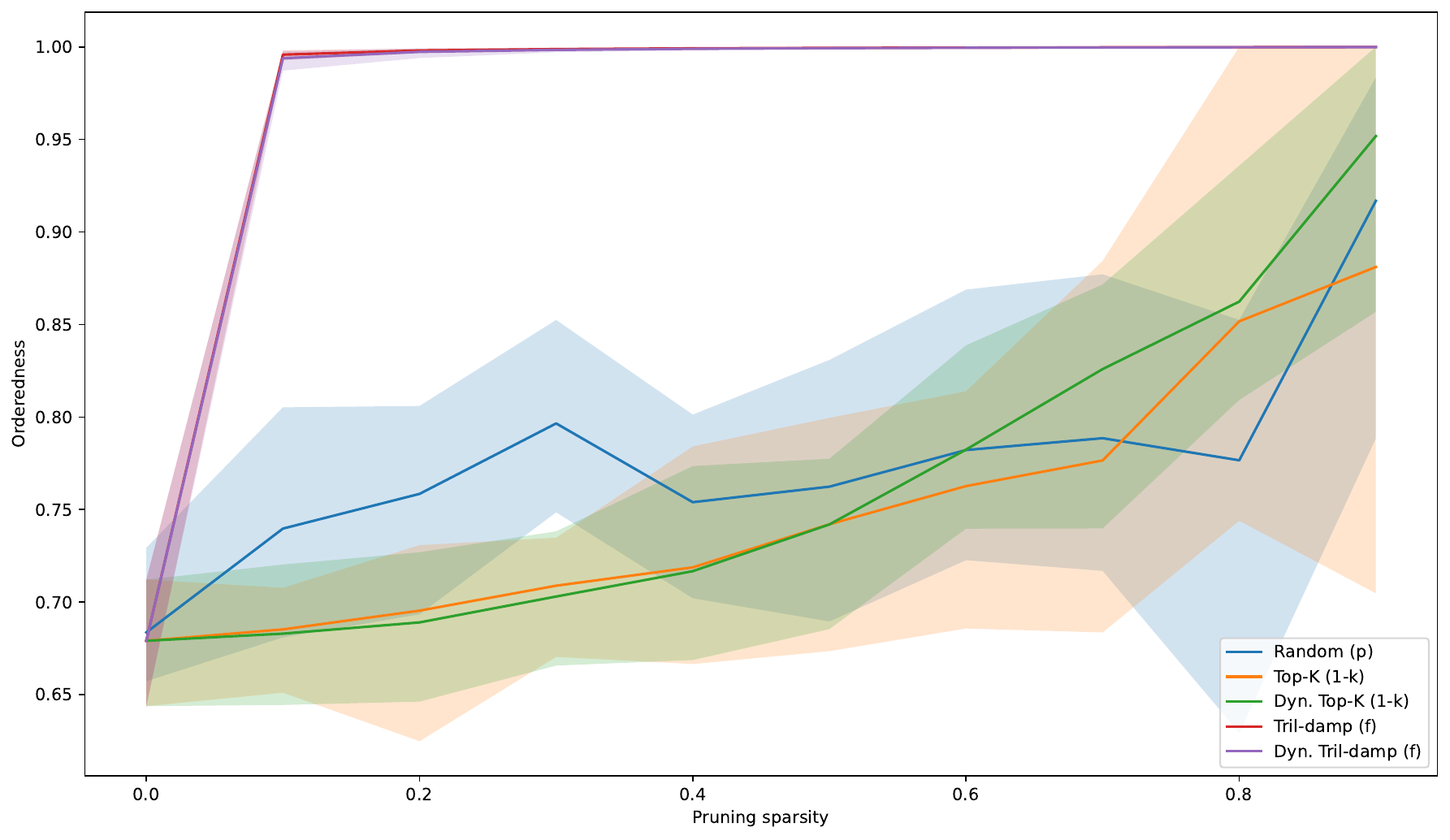}
    \caption{Relationship between pruning sparsity \\ and orderedness (Sine).}
\end{subfigure}
\hfill
\caption{The relationship between target pruning sparsity and orderedness for the XOR and Sine tasks (the measure of increasing sparsity in terms of pruning coefficients is enclosed by brackets in the legend). The equivalent plots for an untrained network are shown for control. Note that since the pruning logic was evolved for only 10 steps on the untrained network, Tril-damping methods were less efficient at damping lower-triangular values at the same sparsity level than the networks trained on XOR and Sine tasks; this difference is therefore not indicative of anything else other than our experimental setup.}
\label{fig:so-plot}
\end{figure}

\paragraph{Effect of Hidden State Size and Evolution Iterations.} As the number of hidden units is increased, the expressive potential of the complete perceptron layer also increases. Therefore, it may be expected that more complicated layering structures can emerge from such networks, provided that the number of evolution iterations also increases. However, as shown in \cref{fig:hi-plot}, this simple prediction is not the case. Based on the data, the relationship between the number of hidden states, evolution iterations, and orderedness is still unclear. However, one striking feature that emerges is the `spike' of orderedness when the network is evolved for 2 iterations only in the XOR task, across all different numbers of hidden units. One possible explanation is that a 2-iteration complete perceptron is structurally highly directional: in the first iteration, the values of the inputs propagate throughout the network; and in the next iteration, the output is extracted. Therefore, any feedback from one hidden unit to another is meaningless, because it either does not utilise the inputs or will never be fed to the output units; at least 3 iterations are required for the input to propagate into the hidden units, be fed back amongst its hidden neurons, and subsequently be used to compute the output.

\paragraph{Sparsity and Orderedness.} \cref{fig:so-plot} gives a closer analysis of how pruning correlates with orderedness. There does appear to be some trend for both tasks, where orderedness increases with sparsity for Top-K methods on top of the levels predicted by the untrained control. However, the variance of the data is too large for any substantial conclusions to be drawn. More theoretical and empirical work is required to understand precisely how sparsity affects orderedness.

\section{Conclusion}

We introduce the complete perceptron layer and {\it orderedness} as a metric for directionality in neural network computation. Through experiments on the complete perceptron layer with two synthetic tasks, we show that orderedness can be induced rather than hard-wired by applying appropriate pruning techniques in conjunction with gradient descent.

\section*{Software and Data}

The code to reproduce this work is available at the following GitHub repository: \url{https://github.com/PerceptronV/orderedness-by-pruning}.

\section*{Acknowledgements}

We are grateful to Kazuki Irie and Hanming Ye for helpful, spontaneous discussions. We also thank Gabriel Kreiman and Giordano Ramos-Traslosheros for their comments and, respectively, teaching and TA-ing the course that inspired this work.

\bibliography{references}
\bibliographystyle{icml2025}

\appendix

\section{Algorithmic Specification of a Complete Perceptron Layer}

\begin{algorithm}[!htbp]
\caption{Forward Computation of a complete perceptron layer}
\label{alg:complete}
\begin{algorithmic}[1]
\REQUIRE Input $\vec{x} \in \R^{B \times i}$, parameters $(\mat{W}, \vec{v}, \vec{b})$, iterations $T$, element-wise activation function $\sigma$
\STATE $\vec{s} \leftarrow \text{repeat}(\vec{v},B)$ \COMMENT{stack one copy of $\vec{v}$ for each batch}
\FOR{$t=1$ to $T$}
    \STATE $\vec{h} \leftarrow [\vec{s}\,\,\vec{x}]$ \COMMENT{append clamped input}
    \STATE $\vec{s} \leftarrow \sigma\big(\vec{h}\,\mat{W}^{\top} + \vec{b}\big)$
    \COMMENT{let information propagate through network}
\ENDFOR
\STATE Return $\vec{s}_{[:,1:o]}$ \COMMENT{first $o$ units constitute the layer output}
\end{algorithmic}
\end{algorithm}

\section{Initialisation and Pruning Strategies}
\label{sec:init-prune}

\paragraph{Initialisation schemes.} \textit{Random Normal} samples elements from $\mathcal{N}(0,1)$; it is the default initialisation method for the weights matrix $\mat{W}$ and values vector $\vec{v}$. \textit{Random Uniform} samples elements from $\mathcal{U}(0,1)$; it helps establish whether the lack of negative initialisation affects orderedness. \textit{Zeros} sets every element to be $0$.

\paragraph{Pruning operations.} \textit{Random Prune} independently zeros each weight with probability $p$.
\textit{Top-K Prune} keeps the top-$k$ fraction of weights by absolute value and zeros the rest. \textit{Dynamic Top-K} does not aggressively select for the top-$k$ fraction right from the start, but instead begins by keeping a much larger fraction $k'$ of the weights, with $k'$ gradually approaching $k$ as training progresses. The function we use is the following:
\begin{equation}
    k'(x)=1-\bigl(1-k\bigr)\sin^{4}\!\Bigl(\tfrac{\pi}{2}x\Bigr), \quad x\in[0,1],
\end{equation}
where $x$ is the training progress. \textit{Tril-Damping} subtracts a fixed fraction of each value from the lower triangle of the weight matrix:
\begin{equation}
    \mat{W}\leftarrow \mat{W} - f\,\mathrm{tril}(\mat{W},-1)
\end{equation} where $f$ is a factor between $0$ and $1$. \textit{Dynamic Tril-Damping} follows a similar pattern to Dynamic Top-K pruning, but instead it starts with a small factor $f'$ so that the lower triangle is not aggressively dampened at the start of training, with $f'$ approaching $f$ as training progresses. The function we use is the following:
\begin{equation}
    f'(x)=f\,\sin^{4}\!\Bigl(\tfrac{\pi}{2}x\Bigr),  \quad x\in[0,1].
\end{equation}

\section{Hyperparameters}
\label{sec:hyperparam}

For the XOR task, by default we used a complete layer with 5 hidden units, an evolution time of 3 iterations, a batch size of 4, and trained the model for 1000 steps (which was found to be sufficient during initial hyperparameter tuning); for the Sine task, the default was 10 hidden units, an evolution time of 3 iterations, a batch size of 10, and training time of 600 steps. For results on an untrained model, we used the same defaults as the XOR task but only evolved our pruning logic for 10 steps, so for low values of $f$, Tril-damping methods were less efficient at inducing orderedness (as can be seen in \cref{fig:so-none}). Across all tasks, the sigmoid activation function was applied and random uniform distribution was applied on the weights matrix and values vector; bias was not used. In all experiments, we employed the Mean Squared Error loss function with the Adam optimiser \citep{kingma2014adam} at a learning rate of 0.01, and ran each setup with 10 different seeds for reliability.

\section{Training Loss of MLP and Complete Perceptron Layers}
\label{sec:training}

The following plots show the training losses of the complete perceptron layer and a traditional MLP on the XOR and Sine tasks. It confirms that the complete perceptron layer, even when pruned during training, is capable of learning.

For the MLP, we used one hidden layer with 2 units on the XOR task, and one hidden layer with 10 units on the Sine task. Bias, Sigmoid activation, and early stopping (after training loss decreases past $10^{-3}$) were used for both tasks. The batch size, optimiser, and learning rate are the same as those of the default complete perceptron layer settings.

For the complete perceptron layers, default settings are used with DynamicTopK pruning ($k$=0.5).

\begin{figure}[!h]
\centering
\hfill
\begin{subfigure}{0.49\textwidth}
    \centering
    \includegraphics[width=\textwidth]{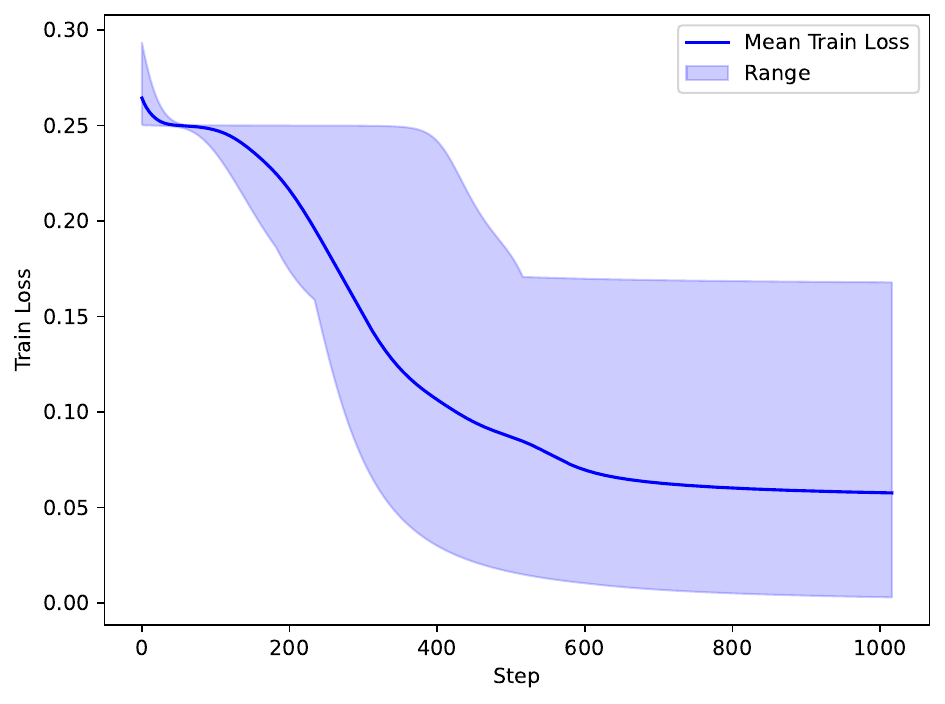}
    \caption{Training loss of a normal MLP for the XOR task.}
\end{subfigure}
\hfill
\begin{subfigure}{0.49\textwidth}
    \centering
    \includegraphics[width=\textwidth]{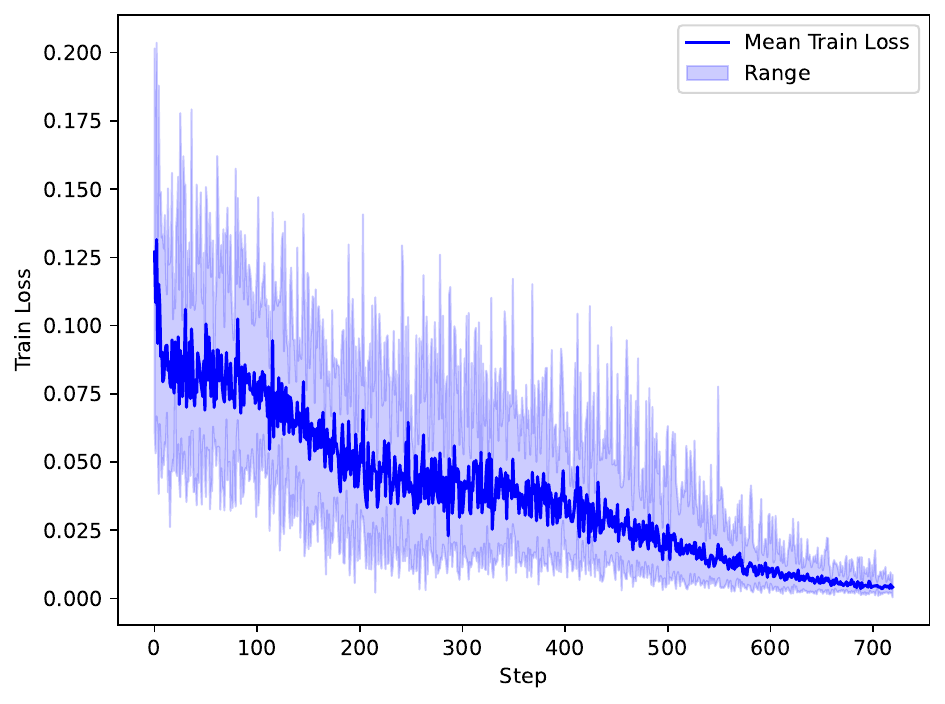}
    \caption{Training loss of a normal MLP for the Sine task.}
\end{subfigure}
\hfill
\caption{Training loss of a normal MLP for the XOR and Sine tasks.}
\label{fig:mlp-dynamics}
\end{figure}

\begin{figure}[!h]
\centering
\hfill
\begin{subfigure}{0.49\textwidth}
    \centering
    \includegraphics[width=\textwidth]{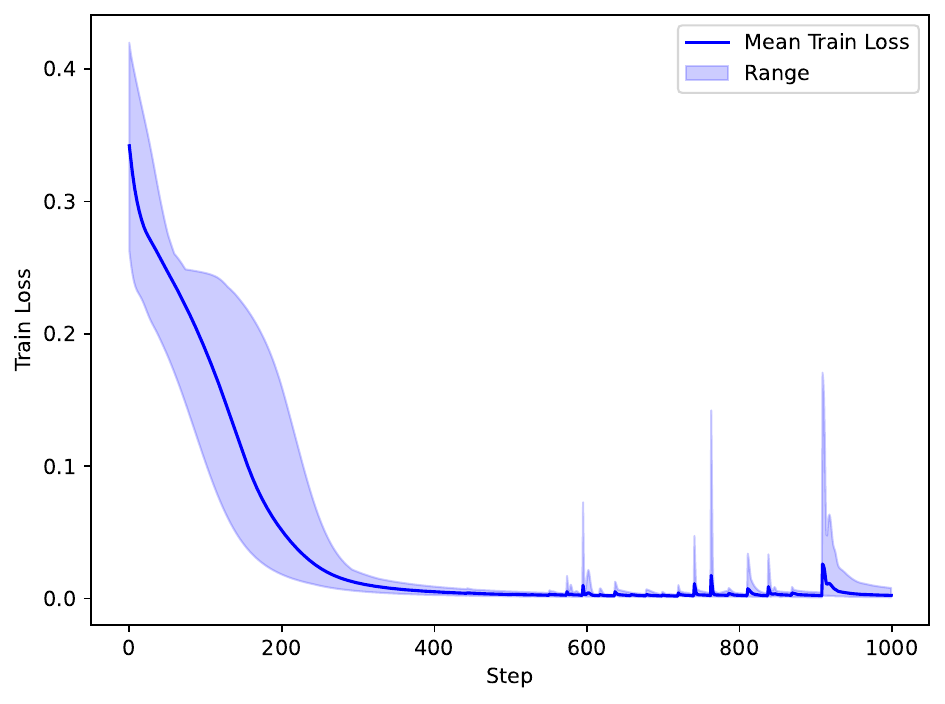}
    \caption{Training loss of a complete perceptron layer for the XOR task; default settings are used with DynamicTopK pruning ($k=0.5$).}
\end{subfigure}
\hfill
\begin{subfigure}{0.49\textwidth}
    \centering
    \includegraphics[width=\textwidth]{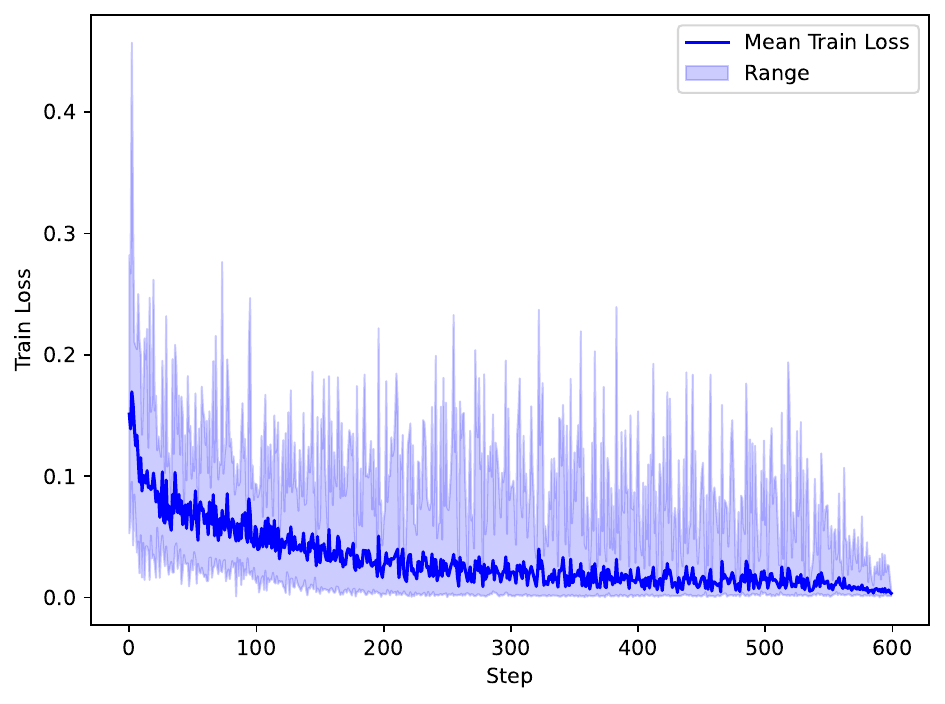}
    \caption{Training loss of a complete perceptron layer for the Sine task; default settings are used with DynamicTopK pruning ($k=0.5$).}
\end{subfigure}
\hfill
\caption{Training loss of a complete perceptron layer for the XOR and Sine tasks.}
\label{fig:dynamics}
\end{figure}


\end{document}